\documentclass[10pt,twocolumn,letterpaper]{article}

\usepackage{etoolbox,lipsum}
\usepackage{iccv}
\usepackage{times}
\usepackage{epsfig}
\usepackage{graphicx}
\usepackage{amsmath}
\usepackage{amssymb}
\usepackage{mathtools}
\usepackage{caption}
\usepackage{subcaption}
\usepackage{bbm}
\usepackage{multirow}

\DeclareMathOperator*{\argmin}{argmin}

\usepackage[pagebackref=true,breaklinks=true,letterpaper=true,colorlinks,bookmarks=false]{hyperref}

\iccvfinalcopy 

\def\iccvPaperID{11} 

\ificcvfinal\fi

\begin{document}

\title{Vision-Guided Forecasting - Visual Context for Multi-Horizon Time Series Forecasting}

\author{Eitan Kosman\\
Technion - Israel Institute of Technology\\
{\tt\small eitan.k@cs.technion.ac.il}
\and
Dotan Di Castro\\
Bosch Center for Artificial Intelligence\\
{\tt\small Dotan.DiCastro@bosch.com}
}

\makeatletter
\patchcmd{\@maketitle}{optional}{\myfigure}{}{}
\newcommand\myfigure{%
  \makebox[0pt]{\includegraphics[width=0.8\linewidth,height=1.1in]{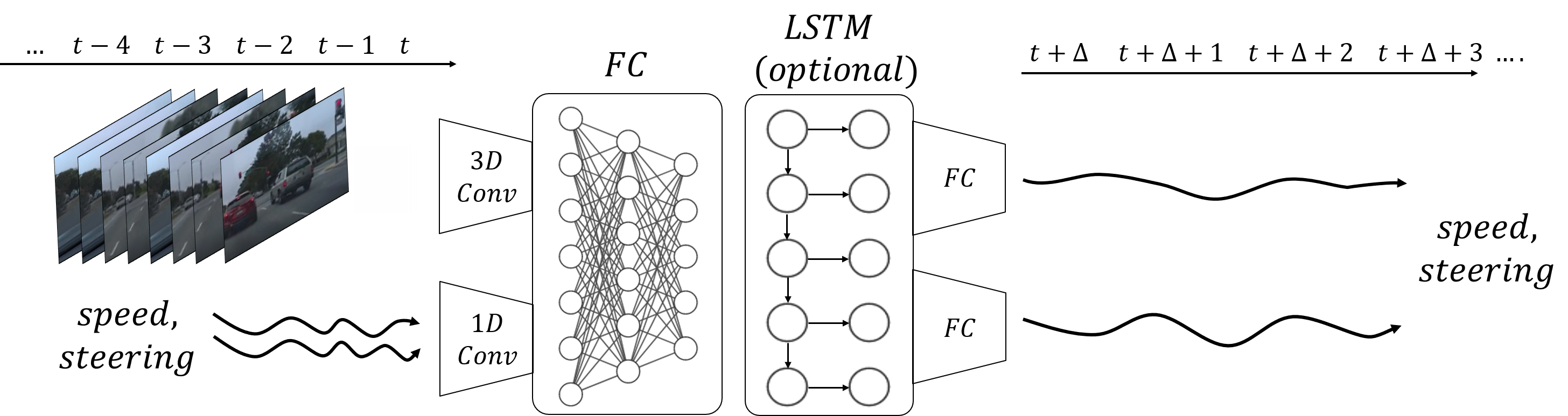}} \\[\normalbaselineskip]
  \refstepcounter{figure}\normalfont{Figure~\thefigure: Problem description. We study the problem of multi-horizon forecasting of driving states from history traces of driving signals, such as steering and speed sensor data, with additional visual information from a front-facing camera.}
  \label{fig:problem_def}
}
\makeatother

\maketitle

\ificcvfinal\thispagestyle{empty}\fi

\begin{abstract}
Autonomous driving gained huge traction in recent years, due to its potential to change the way we commute. Much effort has been put into trying to estimate the state of a vehicle. Meanwhile, learning to forecast the state of a vehicle ahead introduces new capabilities, such as predicting dangerous situations. Moreover, forecasting brings new supervision opportunities by learning to predict richer a context, expressed by multiple horizons. Intuitively, a video stream originated from a front facing camera is necessary because it encodes information about the upcoming road. Besides, historical traces of the vehicle's states gives more context. 
In this paper we tackle multi-horizon forecasting of vehicle states by fusing the two modalities. We design and experiment with 3 end-to-end architectures that exploit 3D convolutions for visual features extraction and 1D convolutions for features extraction from speed and steering angle traces. To demonstrate the effectiveness of our method, we perform extensive experiments on two publicly available real-world datasets, Comma2k19 and the Udacity challenge. We show that we are able to forecast a vehicle's state to various horizons, while outperforming the current state of the art results on the related task of driving state estimation. We examine the contribution of vision features, and find that a model fed with vision features achieves an error that is 56.6\% and 66.9\% of the error of a model that doesn't use those features, on the Udacity and Comma2k19 datasets respectively.
\end{abstract}

\section{Introduction}\label{section:introduction}

Daily driving subconsciously involves many simultaneous tasks, each of which is required for keeping a vehicle on its route safely. The fusion of our senses makes it possible to gather knowledge about our world in order to make it possible to miraculously calculate the commands that are later transferred to our movement organs that control the vehicle.

Despite the great difference between human and autonomous driving, implied by our world-experience and prior knowledge, autonomous vehicles are expected to achieve human-level performance or even surpass it. This suggests that an agent should be provided with state information at the level of detail provided by our senses or even more. Moreover, various components cannot be treated individually due to their interdependency, for example, acceleration and steering. Consequently, attempts that claim to estimate steering commands alone from a single-image, without considering other sensors are doomed to failure in unseen real world scenarios.

\begin{figure}[ht]
\centering
\begin{minipage}{.5\columnwidth}
\centering
\includegraphics[width=\textwidth]{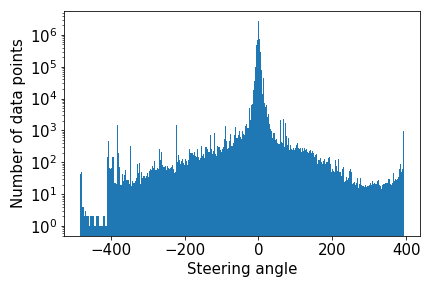}
\end{minipage}\hfill
\begin{minipage}{.5\columnwidth}
\centering
\includegraphics[width=\textwidth]{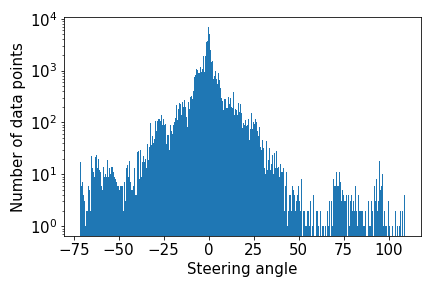}
\end{minipage}
\caption{Histogram of the steering angles in log scale, represented as the double long-tailed distribution; Left - comma.ai. Right - Udacity.}
\label{fig:steer_bias}
\end{figure}

\begin{figure}[ht]
\centering
\begin{minipage}{.5\columnwidth}
\centering
\includegraphics[width=\textwidth]{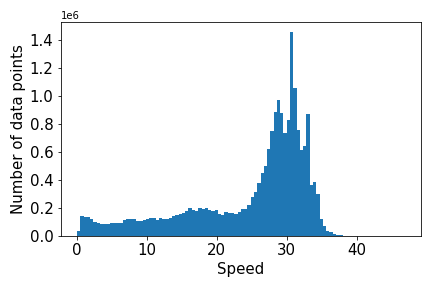}
\end{minipage}\hfill
\begin{minipage}{.5\columnwidth}
\centering
\includegraphics[width=\textwidth]{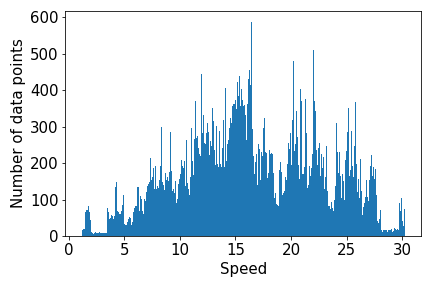}
\end{minipage}
\caption{Histogram of the speed values for the comma.ai dataset (left) and Udacity dataset (right).}
\label{fig:speed_bias}
\end{figure}

Only a few methods pay attention to the distribution of the driving signals. Specifically, the angle of the steering wheel is most of the time kept low during most driving scenarios. As a result, this makes extreme steering angles very rare in the training data, as depicted in Figure \ref{fig:steer_bias}. In contrast, this cannot be claimed for speed values, since they are relatively evenly distributed, as depicted in Figure \ref{fig:speed_bias}. To address this phenomenon, Yuan \etal proposed \emph{SteeringLoss} \cite{yuan2019steeringloss, yuan2020steeringloss} to tackle this problem by designing weighted loss functions. Other methods, such as \cite{bojarski2016end}, attempt to enrich the data using augmentation techniques. However, mimicking real-world dynamics is extremely difficult when considering the representation as a fusion of various signals, such as camera frames, speed, and steering values. Even though, the bias towards certain values requires attention during evaluation because it causes a biased predictor perform artificially well.

Another aspect of reliable control for autonomous vehicles is the ability to provide control commands in real-time. Advances in computing system for autonomous vehicles have taken place in recent years \cite{9251973, collin2020autonomous, bacchus2020accuracy}. However, it is likely that additional requirements in the future will pose additional difficulties. Thus, we want to examine the feasibility of predicting the states ahead of time. By forecasting, we deliver real-time performance and open the door to various uses of these predictions, e.g., anomaly detection. Additionally, forecasting brings the benefits of self-supervision \cite{mao2020survey} and multi-task learning \cite{ruder2017overview} via prediction of multiple time points into the future.

The motivation for forecasting has more natural origins. First, the average response time of a human driver is greater than half a second \cite{jureki_rafal}. Second, most of the driver's attention is focused on the front visual view, consisting of the route that the vehicle will pass thru in the immediate future timestamps. This suggests that driving decisions are not made instantaneously but rather done by short time planning. This makes it highly valuable to predict the future state of the vehicle given the current measurements. Moreover, by forecasting we can avoid dangerous situations ahead of time, by treating scenarios where anomalous behavior is predicted.

To summarize, the task tackled in this paper is learning to predict an ego-vehicle's future states by fusing multiple data modalities: historical traces of speed and steering, and camera frames recorded from a front-facing camera that observes the upcoming track. Although much research has been done on estimating the current state of a vehicle \cite{gidado2020survey}, our goal is different because we focus on forecasting rather than estimating the current state. The setting of the problem is illustrated in Figure \ref{fig:problem_def}.
The main contributions of this paper are as follows:
\begin{enumerate}
    \item We present the problem of vehicle state forecasting as a fusion of multiple signal sources. 
    \item We exploit 3 architectures for multi-horizon forecasting of vehicle signal values. To the best of our knowledge, we are the first to experiment with multi-horizon forecasting in the context of autonomous driving.
    \item We test the hypothesis whether visual information can boost up the performance of predicting a future vehicle's state.
    \item We propose a new protocol for evaluating the performance of prediction models with highly imbalanced datasets in the context of driving, \eg, steering angle and speed prediction of the car.
    \item We present an extensive experimental evaluation of our models. Specifically, we demonstrate the benefits of simultaneous multi-horizon forecasting over single timestamp prediction.
\end{enumerate}

\section{Related Work}\label{section:related_work}

Deep learning for autonomous driving \cite{grigorescu2020survey} was observed as early as 1989, when Pomerleau \etal proposed ALVINN \cite{pomerleau1989alvinn}, a 3-layer neural network for processing images and laser range finder signal, in order to predict the direction in which the vehicle should travel.
The Renaissance of deep learning \cite{lecun2015deep} has given a back support for developing this field further, renewing this by exploiting the power of Convolutional Neural Networks (CNN) for visual understanding. Bojarski \etal~\cite{bojarski2016end} proposed to predict steering angles with only three front-view cameras and manage to control the vehicle. They demonstrated the success of their approach in relatively simple scenarios while being simple to train and deploy. Du \etal~\cite{du2019self} proposed more sophisticated models to perform steering angle prediction using 3D-CNN \cite{mittal2021survey} and \textit{Long Short-Term Memory} (LSTM) modules \cite{hochreiter1997long}. State-of-the-art performance was achieved in the Udacity challenge \cite{udacity} which aimed to predict steering angle based on images only. Besides advances in neural network architectures, ad-hoc objectives were developed, \ie, loss functions. Yuan \etal~ introduced \textit{SteeringLoss} \cite{yuan2019steeringloss, yuan2020steeringloss}, a cost-sensitive loss function, in order to cope with the imbalanced distribution of the steering values.

Another line of works incorporates multi-task learning \cite{ruder2017overview}, side-task learning, or auxiliary task learning, in order to enhance the performance of a model on its main task. Xu \etal~ \cite{xu2017endtoend} use semantic segmentation as the extra supervision and show that this strategy improves performance, especially when coercing a model to attend to small relevant scene phenomena. Hou \etal~ \cite{hou2019learning} take a different approach of enriching the context of the data by distilling knowledge from multiple heterogeneous auxiliary networks that perform related tasks such as image segmentation \cite{minaee2020image} or optical flow \cite{tu2019survey}. While being beneficial for the context enrichment issue, their method does not require additional expensive annotations for the related tasks, which is made possible by using off-the-shelf networks and mimicking their features in different layers. In the context of autonomous driving, in which every vehicle is equipped with a tremendous amount of sensors, it is also reasonable to enrich the context of every state by learning to predict many signals simultaneously, \eg steering angle and speed control as proposed by Yang \etal~\cite{yang2018end}. 

The abundance of sensing devices has opened the door to incorporating richer observations as well. Consequently, approaches relying on multi-modality gained attention recently \cite{xiao2020multimodal, johnson2020feudal}. Xiao \etal~\cite{xiao2020multimodal} analyzed the combination of RGB and depth data (produced by RGBD cameras) based on various fusion schemes, \eg~, early, mid, and late fusion, and showed that the data fusion outperforms single-modality. Johnson and Dana \cite{johnson2020feudal} took an original approach for fusing data from dash-cam images with steering, braking, and throttle signals. They construct a hierarchical framework consisting of manager-worker networks, where the manager, which is fed with historical steering values, gives a high level goal to the worker, which is fed with a sequence of images and intended to give accurate and robust predictions.

In contrast to the previous works, we focus on forecasting future values of the signals from a fusion of video, steering and speed data. To the best of our knowledge, little or no research has been done on the topic. Moreover, we utilize sensor values of multiple timestamps in order to perform multi-task learning, which is a technique that has never been used before in the context of autonomous driving. 

\section{Background}\label{section:background}
\paragraph{Video processing with 3D-CNN's.} 
Video understanding has attracted computer vision researchers for decades. While many computer vision algorithms focus on image-based analysis, video processing introduces the temporal relation between images which raises new difficulties. Consequently, most contributions in this field focus on extending image-based algorithms with the temporal dimension. Examples are \textit{SIFT-3D} \cite{scovanner20073}, \textit{HOG3D} \cite{klaser2008spatio}, \textit{Action-Bank} \cite{sadanand2012action}, and adaptions of convolutional neural networks \cite{mittal2021survey, tran2018closer, yue2015beyond}. Attempts to adapt the latter include LSTM-based network that firstly perform per-frame encoding \cite{srivastava2015unsupervised, ullah2017action, yue2015beyond}. Another line of works rely on the adaptation of the 2D convolution operators themselves to the temporal dimension, \eg, \emph{3D\ CNN} \cite{ji20123d, tran2015learning}, \textit{R2D} \cite{simonyan2014two}, \textit{R(2+1)D} \cite{tran2018closer}, and many more.

\begin{figure}[!h]
    \centering
    \includegraphics[width=0.8\linewidth]{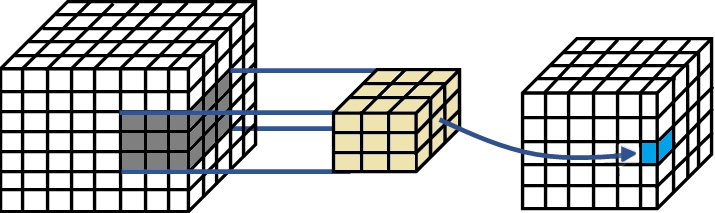}
    \caption{Demonstration of the 3D convolution operator.}
    \label{fig:3d_conv}
\end{figure}

While $3D$ convolutions answer most needs expected from a video processing mechanism, there are still many possibilities to combine it within a neural network. Therefore, we present here a few. For this discussion, \text{$v\in~\mathbbm{R}^{3\times~L\times~H\times~W}$} represents a video clip of $L$ frames, each of which has 3 $RGB$ channels, width of $W$ pixels and height of $H$ pixels. 
In \emph{R2D}, \cite{simonyan2014two}, Simonyan \etal exploit 2D convolutions, which are comprehensively used in machine learning, by ignoring the temporal dimension and reshaping $v$ into a tensor \text{$v'\in~\mathbbm{R}^{3L\times~H~\times~W}$}. Since each filter yields a single channel as output, the first layer of such architecture entirely collapses the temporal information, preventing temporal reasoning in subsequent layers. Other \emph{$3D$ CNN's} models, such as \emph{I3D} \cite{carreira2017quo} and \emph{C3D} \cite{tran2015learning}, relate to architectures whose convolutions are $3D$ kernels only. This is achieved by convolving a 3D kernel with a set of video frames stacked together, resulting with features maps connected to multiple frames in the input, formally denoted in \cite{ji20123d} as (excluding bias and an activation function for ease of exposition)
\begin{equation}
    v_{i,j}^{x,y,z}= \sum_{m}{\sum_{p=0}^{P_i-1}{\sum_{q=0}^{Q_i-1}{\sum_{r=0}^{R_i-1}{w_{i,j,m}^{p,q,r}v_{(i-1)m}^{(x+p)(y+p)(z+r)}}}}}.
\end{equation}
\noindent Here, $(x,y,z)$ is the position on the $j^{th}$ feature map in the $i^{th}$ layer, $R_i$ is the size of the $3D$ kernel along the temporal dimension, and $w_{i,j,m}^{p,q,r}$ is the value of the convolution kernel in the entry corresponding to the coordinates $(p,q,r)$ connected to the $m^{th}$ feature map in the previous layer. This is depicted in Figure \ref{fig:3d_conv}.
Other variants are \textit{MCx} and \textit{rMCx} which are architectures composed of a mixture of $3D$ and $2D$ convolutions. In \textit{MCx} the first layers are $3D$ convolutions followed by $2D$ convolution layers, while in \textit{rMCx} the first layers are $2D$ convolutions followed by $3D$ convolution layers. To this end, the separability of the convolution operator makes it possible to simulate $3D$ convolutions with $2D$ convolutions followed by a $1D$ convolution. One architecture, known as \textit{R(2+1)D} \cite{tran2018closer} makes use of it by decomposing spatial and temporal modeling into two separate steps. Separating the two also has the benefit of adding non-linearities between the two steps. It is also claimed in \cite{tran2018closer} that separating spatial and temporal components renders the optimization easier. The main obstacle of $3D$ models is the enormous number of parameters introduced by $3D$ convolutions, which causes these models to hardly fit into current hardware. Recently, Chen \etal proposed \textit{multi-fiber networks} \cite{chen2018multifiber} for dealing with this issue by slicing a large model into an ensemble of light-weight models, called fibers.

\paragraph{Time series forecasting with deep learning.}  Time series forecasting \cite{mahalakshmi2016survey} relates to the process of using a model to predict future values based on history traces of observed values. It has a vast amount of applications, such as stock market prediction \cite{rao2015survey}, weather forecasting \cite{campbell2005weather}, healthcare prediction \cite{kaushik2020ai}, and many more. Thereof, the simplest case of a forecasting mechanism consists of a function $f:\mathbbm{R}^{p}\to \mathbbm{R}$ where the mapping takes $p$ consecutive values as input and maps values drawn from a time-series $\{x_0,x_1,...\}$ by
\begin{equation}\label{simple_forecsting}
    x_{t+\tau} = f([x_{t-p+1},...,x_{t}]^T).
\end{equation}
In the early days, methods mainly relied on parametric models, such as autoregressive, exponential smoothing, structural time series models, autoregressive integrated moving average (ARIMA) which generalizes the former, and many more. Recently, the availability of massive data and scalable deep-learning frameworks have driven the adaption of neural network architectures for those tasks \cite{lim2020time}, \eg Recurrent neural networks (RNN's) \cite{salinas2020deepar, rangapuram2018deep, lim2019recurrent, wang2019deep} which are powerful to capture non-linear patterns. However, early models suffered from limitations in learning long-range dependencies due to issues with exploding and vanishing gradients \cite{goodfellow2016deep}. \emph{Long Short-Term Memory networks (LSTMs)} \cite{hochreiter1997long} was developed to overcome those limitations by introducing a \textit{cell-state} in order to store long-term information, modulated through a series of gates.

Multi-horizon forecasting relates to the task of estimating at multiple timestamps in the future \cite{patton2011predictability, capistran2010multi}. This generalizes Equation \ref{simple_forecsting}, as the new mapping $f:\mathbbm{R}^p\to~\mathbbm{R}^{m}$ predicts $m$ values ahead:
\begin{equation}
\label{eq:multi-horizon}
    [x_{t+\tau},...,x_{t+\tau +m-1}]^T = f([x_{t-p+1},...,x_{t}]^T).
\end{equation}

\section{Multi-Horizon Forecasting}\label{section:methodology}
\begin{figure}[!h]
    \centering
    \includegraphics[width=0.9\linewidth]{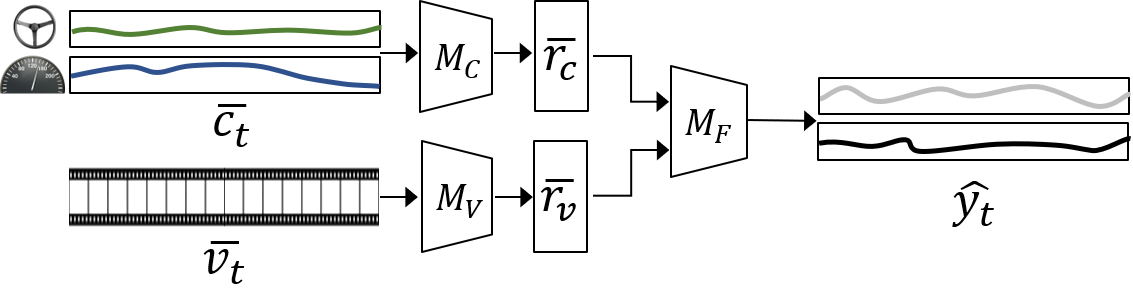}
    \caption{A general scheme of the proposed architecture.}
    \label{fig:framework}
\end{figure}
In this section we formally describe the details of our proposed models. The input is a bi-modal temporal signal  composed of: (1) A sequence of $n$ consecutive frames corresponding to a timestamp $t$, $\bar{v_t}=\{v_{t-n+1},...,v_t\}$, and (2) A sequence of $k_{in}$ values originated from $s_{in}$ sources $\bar{c_t}=\{c_{t-k_{in}+1},...,c_t|c_i\in \mathbbm{R}^{s_{in}}\}$. We allow the two inputs have different lengths and sampling rate, which does not affect the overall design of our system. A multi-horizon forecasting model predicts a sequence of length $k_{out}$ corresponding to consecutive timestamps originated from $s_{out}$ sources, $\hat{y_t}=\{y_{t_1},...,y_{t_{k_{out}}}|y_{t'}\in \mathbbm{R}^{s_{out}}\}$. We rely on a  model consisting of 3 modules: (1) Video Processing Module (VPM; denoted by $M_V$); (2) A Controller Area Network (CAN-Bus) Signals Processing Module (CPM; denoted by $M_C$); (3) A Fusion Module for predicting the output sequences from both feature representations (denoted by $M_F$). See Figure \ref{fig:framework} for an overview of the architecture.

The processing modules, namely $M_C$ and $M_V$, are fed with the inputs $\bar{c_t}$ and $\bar{v_t}$ respectively and encode them to compressed representations $\bar{r_c}=M_C(\bar{c}), \bar{r_v}=M_V(\bar{v})$. The fusion module $M_F$ then performs a multi-horizon forecasting $\hat{y}=M_F(\bar{r_v}, \bar{r_c})$ as described in \eqref{eq:multi-horizon}.

In all of our architectures we keep $M_C$ and $M_V$ unchanged, while we change the structure of $M_F$. In the following sections, we describe the different architectures in detail. In the first architecture, we construct a separate model for predicting signal values for each timestamp, as depicted in Figure \ref{fig:arch_decorr}. In the second architecture, we construct a single model for predicting the output as a whole via a fusion module implemented by a fully-connected layers, as depicted in Figure \ref{fig:arch_corr}. In the third architecture, we use embeddings extracted by the processing modules $M_V$ and $M_C$ as inputs to a LSTM module and predict via a sequence-to-sequence methodology \cite{neubig2017neural}, depicted in Figure \ref{fig:arch_lstm}.

\subsection{Fusion Modules}
\label{module:fusion}
The output of both processing modules are used for the prediction task. Unlike the processing modules, $M_V$ and $M_C$, that remain unchanged across the 3 different architectures, the fusion module is the building block that distinguishes between the various architectures. In the following, descriptions of the three architectures are presented.

\begin{figure}[!h]
    \centering
    \includegraphics[width=0.85\linewidth]{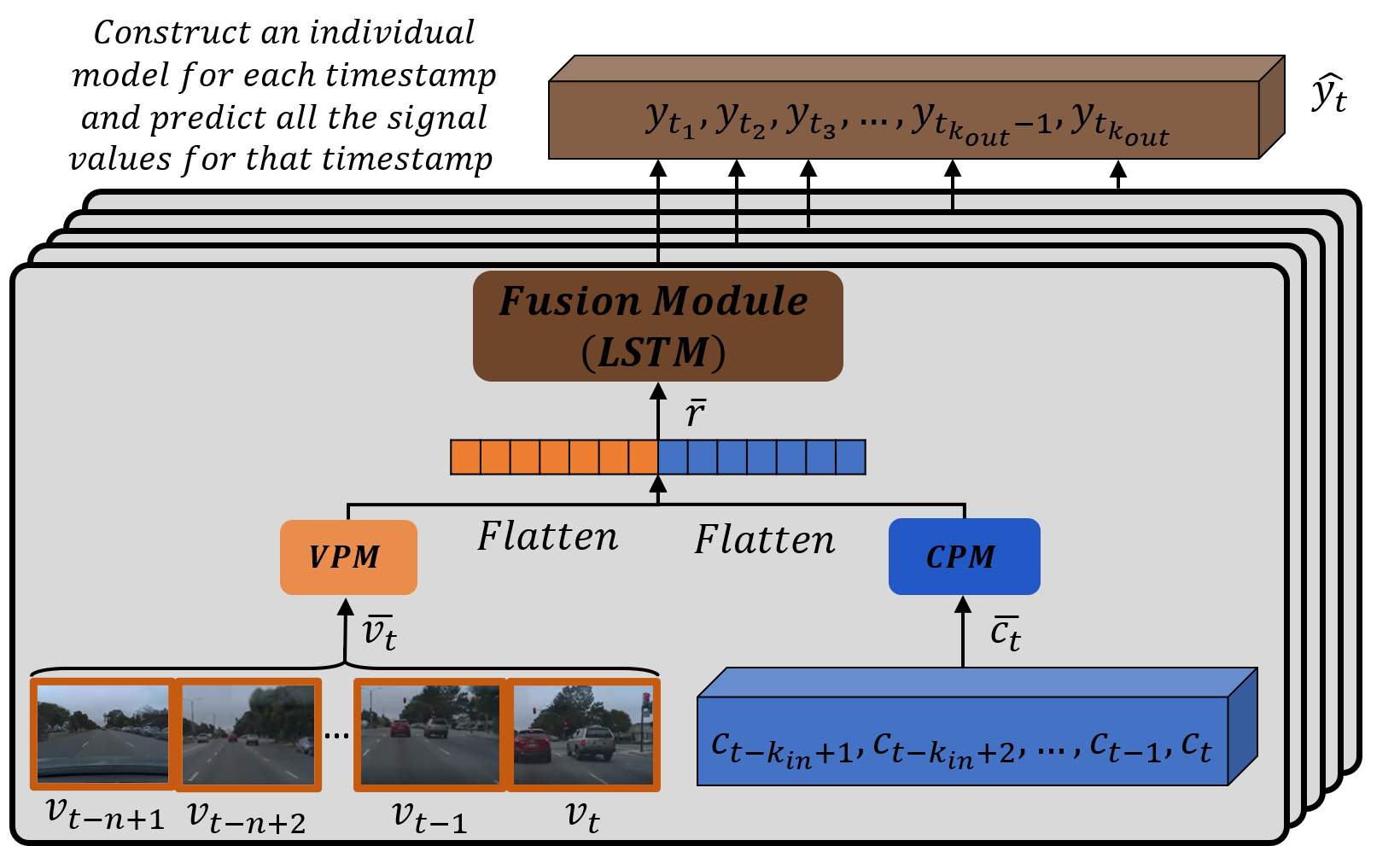}
    \caption{Architecture for individual multi-horizon prediction. We use the fused features from both processing modules to predict signal values corresponding to one timestamp per model.}
    \label{fig:arch_decorr}
\end{figure}

\noindent \textbf{Architecture 1:}
Predicting each timestamp individually with a fully connected fusion module (denoted with MH-IND-FC). The outputs of both $M_V$ and $M_C$ are flattened and concatenated. The result is fed into a fully connected neural network with $Relu$ activations between every two consecutive layers, which outputs predictions for a single timestamp. This architecture is depicted in Figure \ref{fig:arch_decorr}. We duplicate and train this model for each timestamp to perform multi-horizon forecasting.

\begin{figure}[!h]
    \centering
    \includegraphics[width=0.85\linewidth]{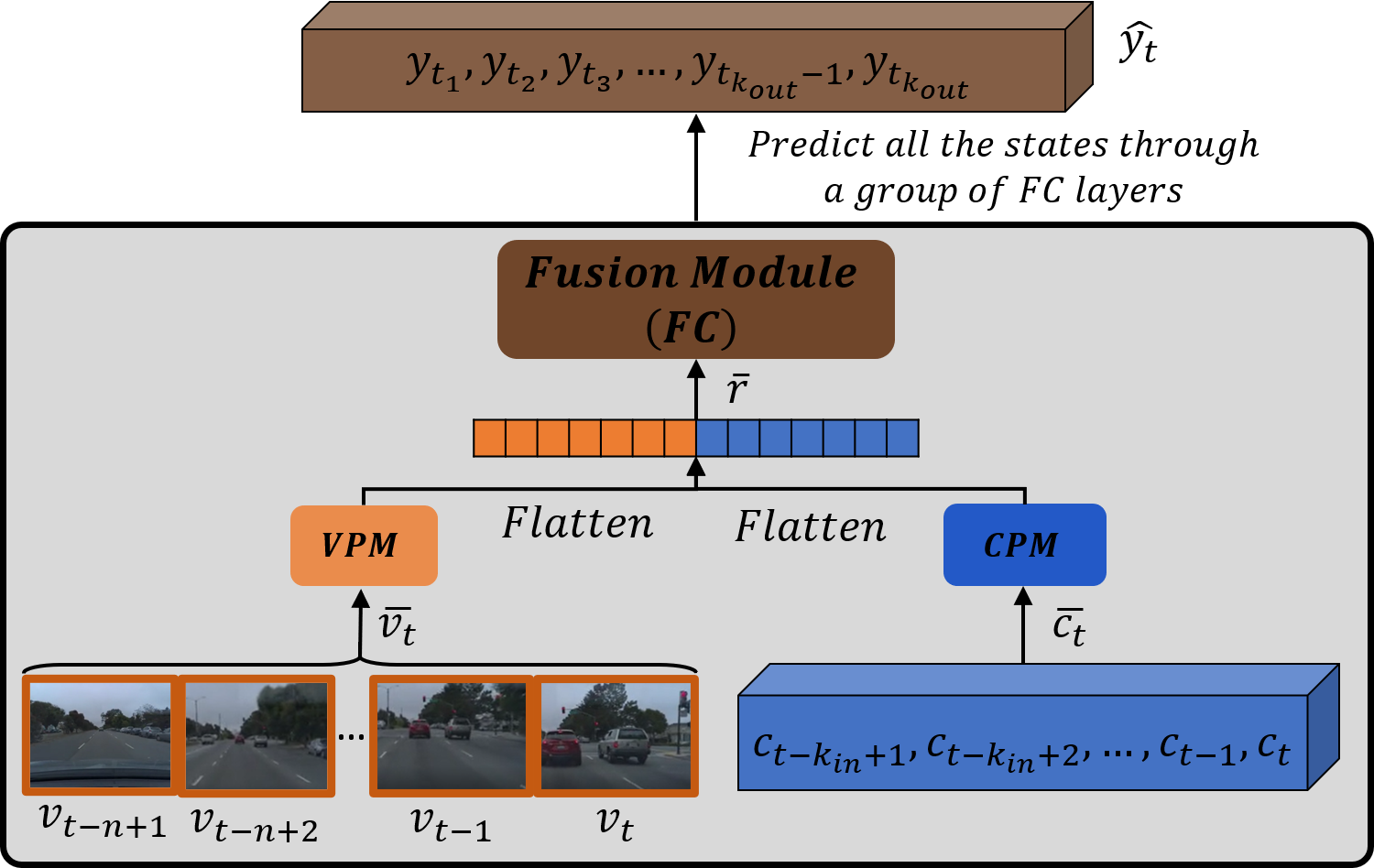}
    \caption{Architecture for simultaneous multi-horizon forecasting. We use the fused features from both processing modules to predict signal values for multiple signals and multiple timestamps simultaneously via a fully-connected based fusion module.}
    \label{fig:arch_corr}
\end{figure}

\noindent\textbf{Architecture 2:} Predicting multi-horizon simultaneously with a fully connected fusion module (denoted with MH-SIM-FC). Similarly, the output of both $M_V$ and $M_C$ are flattened and concatenated. The result is fed into a fully connected neural network with $Relu$ activations between every two consecutive layers, which outputs predictions for all timestamps simultaneously. This architecture is depicted in Figure \ref{fig:arch_corr}. 

\begin{figure}[!h]
    \centering
    \includegraphics[width=0.85\linewidth]{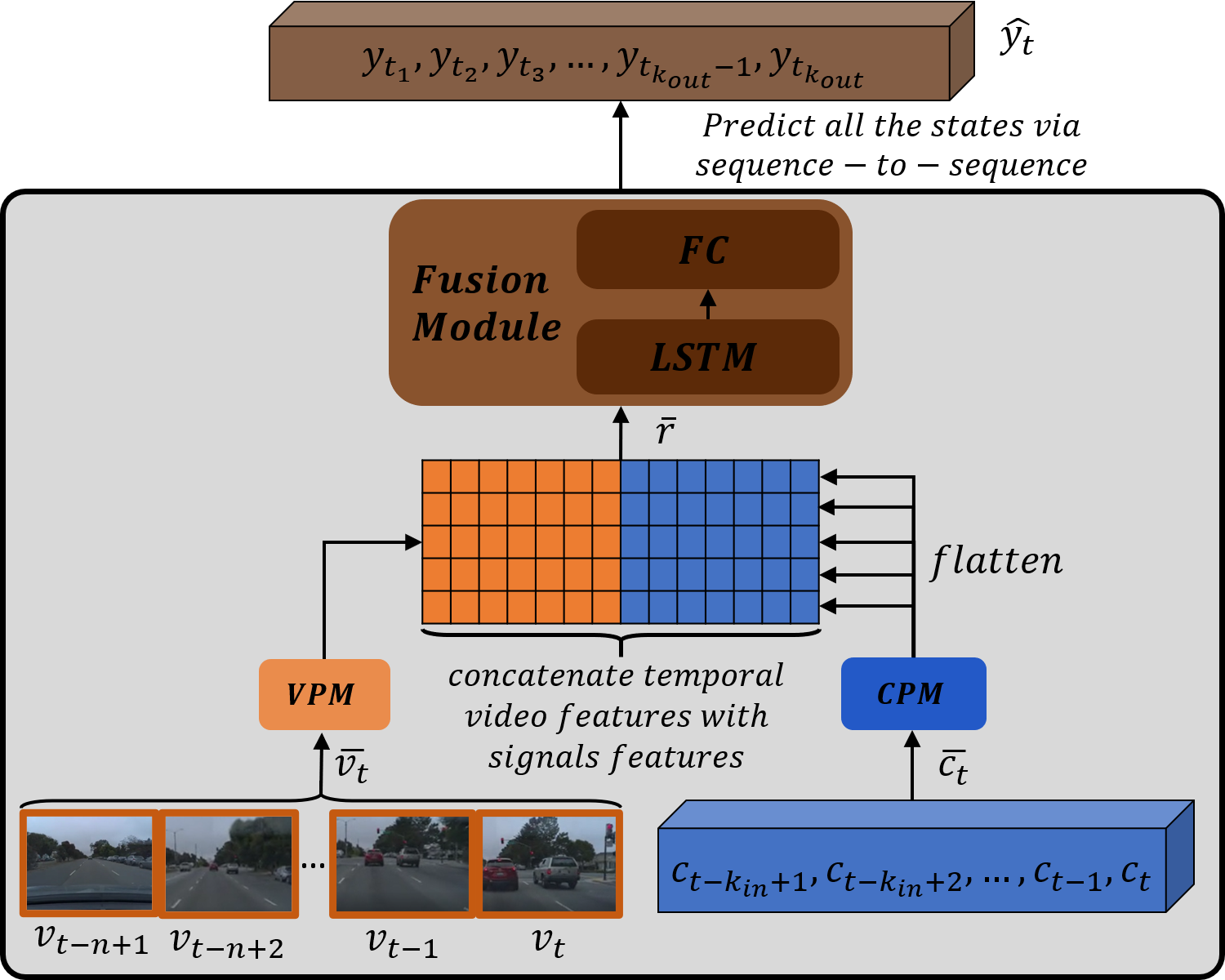}
    \caption{Architecture for simultaneous multi-horizon forecasting. We use the fused features from both processing modules to predict signal values for multiple timestamps together via an LSTM based fusion module.}
    \label{fig:arch_lstm}
\end{figure}

\noindent\textbf{Architecture 3:} Predicting multi-horizon simultaneously via LSTM (denoted with MH-SIM-LSTM). We flatten the output of $M_C$ only and concatenate it with the output of $M_V$ on the dimension corresponding to time, creating a series of feature vectors containing both visual and signals data. The series is fed into an LSTM module followed by a fully-connected decoder for predicting all the values for all timestamps simultaneously. This architecture is depicted in Figure \ref{fig:arch_lstm}.
    
\subsection{Optimization}
Each training sample consists of an input $(\bar{v_t},\bar{c_t})$ and a target $\bar{y_t}$ representing the values of the signals to predict. We denote the predicted variables as $M_F(M_C(c_t),M_V(v_t))$. Our objective is to find the model parameters  $\Theta=\Theta_{M_V}~\cup~\Theta_{M_C}~\cup~\Theta_{M_F}$ that optimizes
\begin{equation}\label{eq:objective}
    \hat{\Theta} = \argmin_{\Theta} {|\bar{y_t} - M_F(M_C(c_t),M_V(v_t))|_{2}^{2}}.
\end{equation}

\section{Evaluation}\label{section:evaluation}
In the following we describe how we evaluate the proposed models.
The conventional evaluation protocol used in the literature is Mean Absolute Error (MAE) and Root Mean Squared Error (RMSE) \cite{hou2019learning, yuan2019steeringloss, yuan2020steeringloss, johnson2020feudal, du2019self, yang2018end}. As comprehensively discussed previously, this protocol does not suit imbalanced datasets. We performed an experiment showing that a biased predictor which outputs a constant value of $0$ achieves a MAE of $4.1$ degrees on \textit{comma.ai} \cite{1812.05752} and $6.9$ on Udacity test set \cite{udacity}. Although these errors are significantly higher than results of previous works that estimate steering angles, it can be considered as negligible because of the relatively high tolerance of the steering wheel, which has a wide range of values anyway. However, the biased predictor completely misses sharp angles, suggesting that MAE poorly quantifies the error. Moreover, a slightly worse error in these measures is forgivable if a predictor is able to perform well in the edge-cases.

We propose evaluating the prediction performance on per-range basis. Given a dataset of $n$ samples where $y_i$ is the target steering angle of the $i^{th}$ sample, and $\hat{y_i}$ is the predicted steering angle, the MAE@$\alpha$ is calculated by
\begin{equation}
    \textrm{MAE@}\alpha = \frac{\sum_{i=1}^{n}{\mathbbm{1}_{|y_i|\geq~\alpha}|\hat{y}_i-y_i|}}{\sum_{i=1}^{n}{\mathbbm{1}_{|y_i|\geq~\alpha}}}.
\end{equation}
With $maxSteering$ as the maximum absolute angle of the steering wheel, we vary $\alpha \in [0, maxSteering]$ and plot the results as a graph. We expect the graph to be monotonically increasing for biased predictors since small angles captured by small values of $\alpha$ has little contribution to the errors sum while increasing the population. In fact, MAE@0 is the standard MAE used in prior work.

Unlike steering angles, speed values are spread uniformly and therefore the conventional evaluation protocols are used.

\begin{table*}[htb]
\small
\centering
\begin{tabular}{ |c|c|c|c|c|c|c|c|c|c| }
\hline
 & & \multicolumn{4}{c|}{Steering} & \multicolumn{4}{c|}{Speed}\\
 \hline
 & & \multicolumn{2}{c|}{comma.ai} & \multicolumn{2}{c|}{Udacity}& \multicolumn{2}{c|}{comma.ai} & \multicolumn{2}{c|}{Udacity}\\
 \hline
 
 Method & Horizon & MAE & RMSE & MAE & RMSE & MAE & RMSE & MAE & RMSE\\
 \hline
 
\multirow{5}{*}{MH-IND-FC} 
 & 0.5 & 1.186 & 4.838 & 1.472 & 1.295  & 0.183 & 0.337 & 2.404 & 2.673\\
 & 1 & 1.742 & 6.874 & 1.591 & 2.119 & 0.179 & 0.315 & 2.531 & 2.739\\
 & 1.5 & 2.228 & 9.18 & 1.468 & 1.959 & 0.259 & 0.439 & 2.751 & 3.053\\
 & 2 & 2.556 & 10.853 & 1.408 & 1.897 & 0.354 & 0.589 & 2.868 & 3.297\\
 & 2.5 & 2.814 & 12.004 & 1.736 & 2.371  & 0.556 & 0.982 & 2.802 & 3.141\\
 \hline
 
 \multirow{5}{*}{MH-SIM-FC}
 & 0.5 & 1.154 & 3.984 & 1.345 & 1.897 & 0.123 & 0.249 & 1.641 & 1.899\\
 & 1 & 1.78 & 7.003 & 1.368 & 1.952 & 0.182 & 0.34 & 1.713 & 2.019\\
 & 1.5 & 2.255 & 9.308 & 1.555 & 2.203 & 0.261 & 0.476 & 1.834 & 2.152\\
 & 2 & 2.552 & 10.568 & 1.592 & 2.233 & 0.352 & 0.617 & 1.929 & 2.294\\
 & 2.5 & 2.757 & 11.473 & 1.761 & 2.456 & 0.446 & 0.769 & 2.102 & 2.41\\
 \hline
 
 \multirow{5}{*}{MH-SIM-LSTM}
 & 0.5 & \textbf{1.08} & \textbf{3.444} & \textbf{0.677} & \textbf{1.394} & \textbf{0.112} & \textbf{0.212} & \textbf{0.133} & \textbf{0.313}\\
 & 1 & \textbf{1.578} & \textbf{6.474} & \textbf{0.772} & \textbf{1.706} & \textbf{0.129} & \textbf{0.335} & \textbf{0.168} & \textbf{0.322}\\
 & 1.5 & \textbf{1.926} & \textbf{8.378} & \textbf{0.781} & \textbf{1.733} & \textbf{0.227} & \textbf{0.415} & \textbf{0.185} & \textbf{0.357}\\
 & 2 & \textbf{2.353} & \textbf{9.657} & \textbf{0.885} & \textbf{1.988} & \textbf{0.301} & \textbf{0.586} & \textbf{0.208} & \textbf{0.395}\\
 & 2.5 & \textbf{2.586} & \textbf{10.442} & \textbf{1.084} & \textbf{2.235} & \textbf{0.388} & \textbf{0.539} & \textbf{0.247} & \textbf{0.462}\\
 \hline
\end{tabular}
\caption{Test results for prediction both steering angle and speed for the multi-horizon forecasting architectures. Values are in degrees. Lower is better. In bold are the best errors for prediction for each column and horizon where is apparent the MH-SIM-LSTM is the best architecture in our case.} 
\label{table:results}
\end{table*}

\section{Experiments}\label{section:experiments}
In this section we evaluate the proposed architectures in several experiments. Specifically, our goals are the following:
\begin{itemize}
    \item Compare the 3 proposed architectures.
    \item Measure the benefits of training with broader context, \ie multi-tasking via multi-horizon forecasting, and conclude if it improves our ability to forecast.
    \item Asses the impact of visual features from video data on forecasting of speed and steering.
\end{itemize}

\subsection{Implementation details}
\paragraph{Architectures and inputs.}
The architectures for our processing modules $M_C$ and $M_V$ are based solely on convolutional operators. 
The module $M_V$ exploits 3D convolutions to extract features from the video clips. Specifically, we adopt the lightweight architecture of \textit{MFNET-3D} \cite{chen2018multifiber} as a backbone and make use of the output of its global average pooling layer for the following steps of the algorithm. This results with a representation \text{$\bar{r_v}\in \mathbbm{R}^{5\times~768}$}. The $M_V$ module is fed with \text{$n=10$} video frames \text{$\bar{v_t}=\{v_{t-n+1},...,v_t|v_i\in~\mathbbm{R}^{3\times~224\times~224}\}$}. With a frame rate of \text{$fps=12.5$}, \text{$\bar{v_t}$} contains visual information from a time window of 0.8 [sec].
The module $M_C$ exploits 1D convolutions to extract features from CAN-bus data originated from sensors mounted on the vehicle, specifically speed and steering angle, represented as a sequence of \text{$k_{in}=10$} states \text{$\bar{c_t}=\{c_{t-k_{in}+1},...,c_{t}|c_i\in~\mathbbm{R}^2\}$}. The signals are sampled at 10 [HZ], which means it contains information from a time window of 1 [sec]. It is implemented as a a 3-layer convolutional neural network with an increasing number of channels $2\to4\to8\to16$ and $Relu$ activations between every two convolutional layers. Each 1D convolution layer is parameterized with a stride of 2 steps in the temporal dimension and zero-padded by one entry. This results with a representation \text{$\bar{r_c}\in~\mathbbm{R}^{32}$}. In the following, the description of the fusion module $M_F$ architecture for the three architectures is given.

\noindent\textbf{MH-IND-FC} (Architecture 1) - Each timestamp has its own forecasting model with individual $CPM$ and $VPM$. Similarly, we flatten and concatenate the two representations to obtain \text{$\bar{r}=[\bar{r_c}^{T},\bar{r_v}^{T}]^T$} and feed with it a fusion module which consists of 2 fully-connected heads implemented by a 3-layer fully connected network with a geometrically decreasing number of neurons \text{$3872\to246\to15\to1$}.\\
\noindent\textbf{MH-SIM-FC} (Architecture 2) - We flatten and concatenate the two representations to obtain $\bar{r}$ and feed with it a fusion module which consists of 2 fully-connected heads, one for each sensor, each of which is implemented by a 3-layer fully connected network with a geometrically decreasing number of neurons \text{$3872\to421\to45\to5$}.\\
\noindent\textbf{MH-SIM-LSTM} (Architecture 3) - We flatten the output of $CPM$ only and concatenate the result with each temporal feature results from $VPM$, \ie \text{$\bar{r}\in \mathbbm{R}^{5\times 800}$}. Each temporal feature of the 5 is fed into a fully connected layer to reduce its dimension to $400$. Next, we feed the results as a sequence consisting of 5 features into an LSTM layer with hidden dimension of 64, followed by an LSTM layer with hidden dimension of 32. The resulting sequence from the last LSTM layer is fed into a 2-layer fully connected with a decreasing number of neurons \text{$32\to16\to2$}.

\paragraph{Data pre-processing, sampling and augmentation.} Data enrichment is achieved by flipping video frames on their horizontal axis and multiplying the corresponding steering values by $(-1)$ with probability of $0.5$. Besides enrichment, this techniques helps dealing with the skew of the steering angle values in both datasets. In addition, we avoid training samples with low speed values since steering angles in these situations are not informative and hurt the training.

\subsection{Datasets}
The first dataset used in our experiment is the \emph{Udacity Driving Dataset} \cite{udacity} which is an open-source collection of video frames captures by a dash cam along with the corresponding steering-angles, braking and throttle pressure data. We use data from Udacity challenge 2, namely CH2\_002. The total time recorded is 1694 seconds in various driving conditions: direct sunlight, lighting changes, shadows, etc. Since there is no pre-defined train/test split, we randomly split it by the ratio of 2:1 for training and testing respectively. The second dataset used in our experiments is the \emph{Comma2k19} \cite{1812.05752} which captures over 33 hours of driving data. It is segmented into 1-minute-long 2019 segments of highway driving between San Jose and San Francisco. It includes frames captured by a road-facing camera, along with phone GPS, thermometers, 9-axis IMU and CAN-bus data. This dataset is not formally divided into train/test splits, thus we randomly select approximately 70\% of the segments for training, and the rest for testing.

\subsection{Results}\label{section:results}
We begin with the analysis of the experimental results of steering angle prediction provided in Table \ref{table:results}. This table shows the error in terms of MAE and RMSE for the 3 architectures described in section \ref{module:fusion}. At the time of writing this paper, state-of-the-art methods achieve an estimation performance 0.091 (radians) on \emph{comma.ai} \cite{yuan2019steeringloss, yuan2020steeringloss} and 0.04659 (radians) on Udacity \cite{johnson2020feudal} in terms of RMSE. Accordingly, the first conclusion is that we were able to predict the steering angle with competitive performance to other state-of-the-art results for the steering estimation task. We also observe that MH-IND-FC achieves the worst performance among the three, suggesting that multi-task learning via multi-horizon forecasting quantitatively improves the results. Moreover, MH-SIM-LSTM  achieves better performance than MH-SIM-FC. This implies that a sequence-to-sequence LSTM as a fusion module is beneficial.

We continue with the analysis of the experimental results of speed prediction provided in Table \ref{table:results}. Similarly to the steering angle forecasting results, MH-IND-FC achieves the worst performance among the three, while  MH-SIM-LSTM achieves the best results. Following the success of the later on both tasks, we conclude that performing multi-task learning through multi-horizon forecasting and incorporating LSTM in the fusion module experimentally improves the performance.

\begin{figure*}[ht]
    \centering
    \begin{subfigure}[b]{0.33\textwidth}
         \centering
         \includegraphics[width=\textwidth]{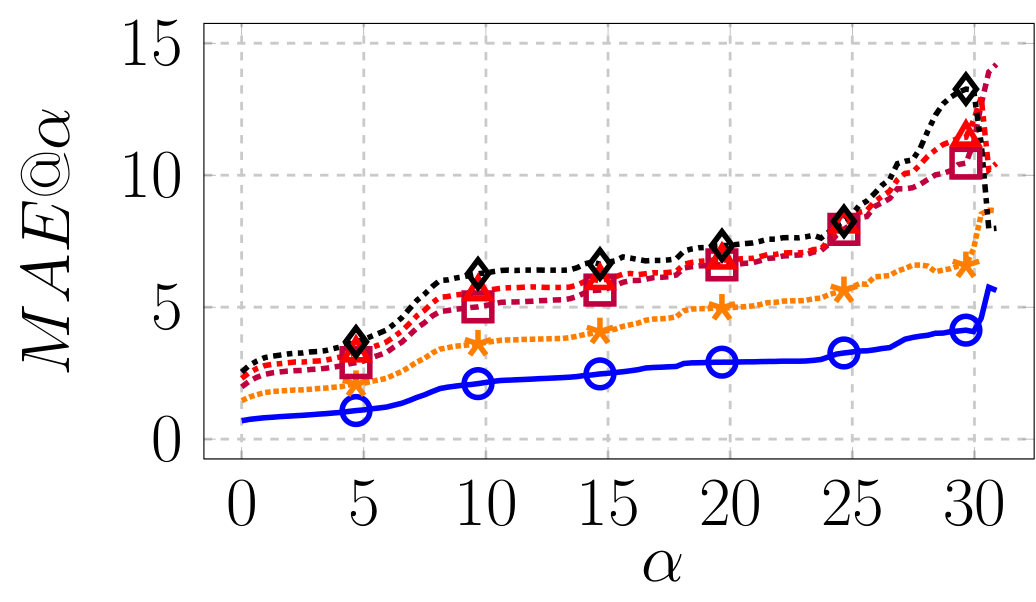}
         \caption{MH-FC-IND (Udacity)}
         \label{fig:mae_mh_ind_fc_udacity}
     \end{subfigure}
    \begin{subfigure}[b]{0.33\textwidth}
         \centering
         \includegraphics[width=\textwidth]{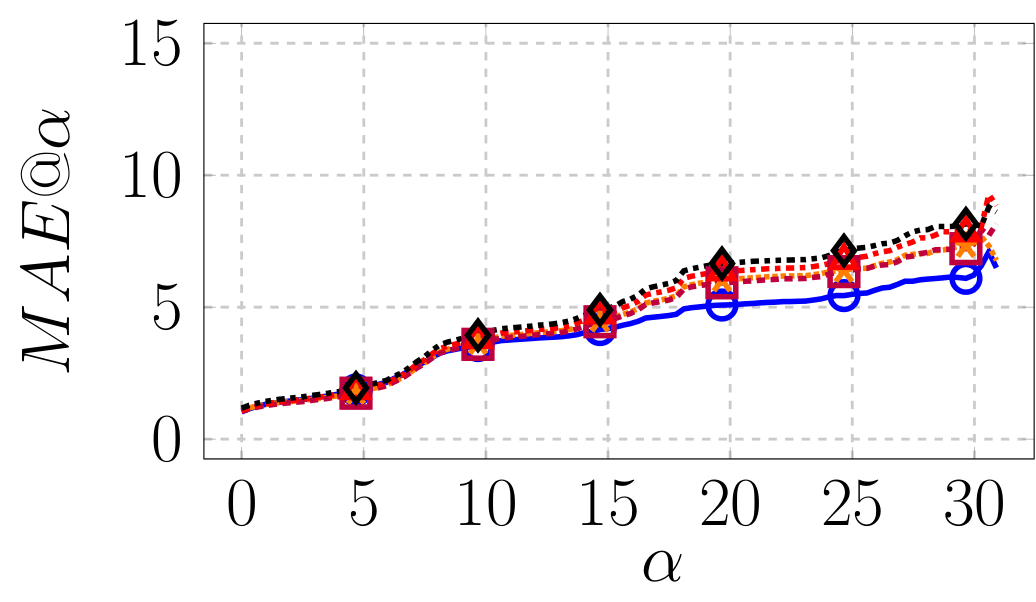}
         \caption{MH-SIM-FC (Udacity)}
         \label{fig:mae_mh_sim_fc_udacity}
     \end{subfigure}
    \begin{subfigure}[b]{0.33\textwidth}
         \centering
         \includegraphics[width=\textwidth]{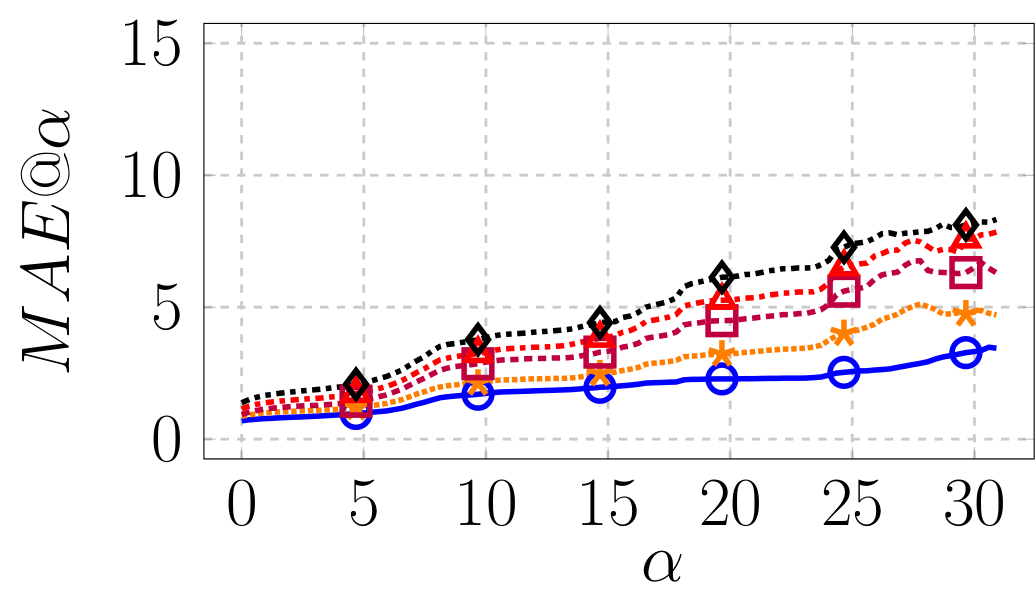}
         \caption{MH-SIM-LSTM (Udacity)}
         \label{fig:mae_mh_sim_lstm_udacity}
     \end{subfigure}

    \begin{subfigure}[b]{0.33\textwidth}
         \centering
         \includegraphics[width=\textwidth]{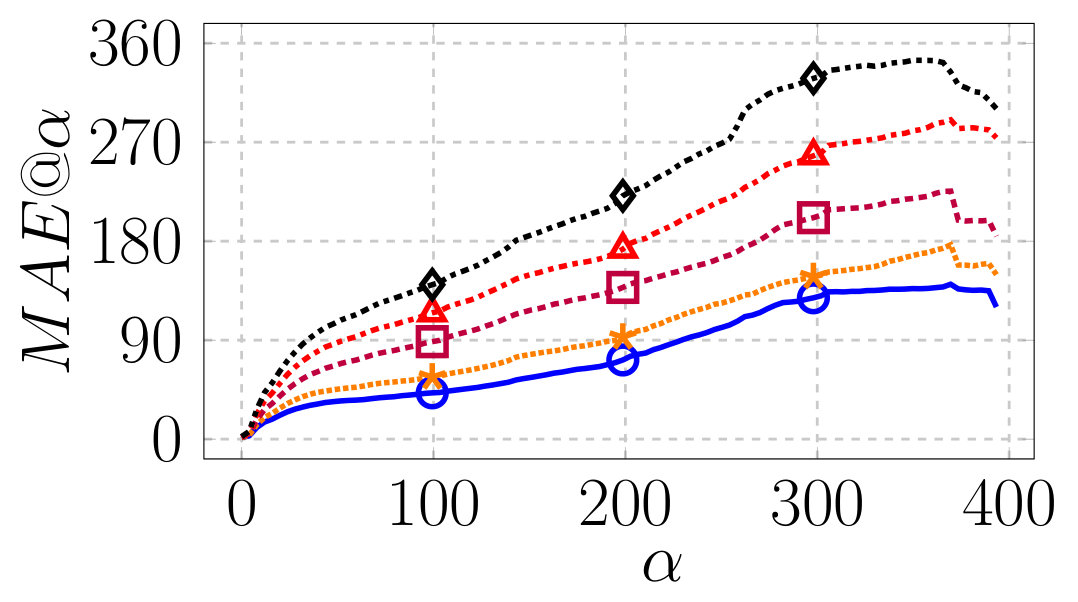}
         \caption{MH-IND-FC (Comma2k19)}
         \label{fig:mae_mh_ind_fc_comma}
     \end{subfigure}
    \begin{subfigure}[b]{0.33\textwidth}
         \centering
         \includegraphics[width=\textwidth]{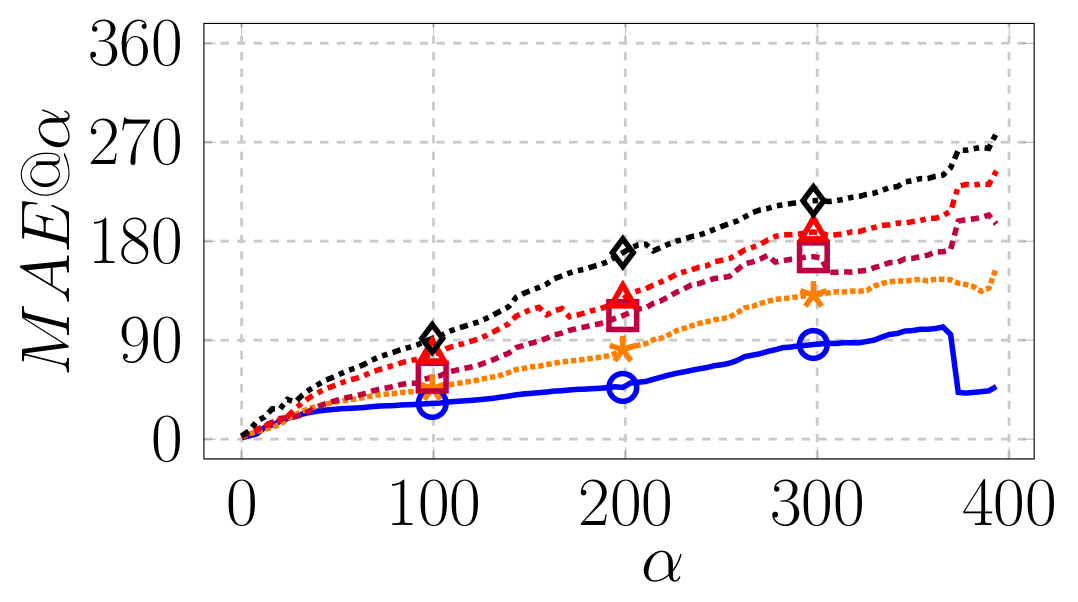}
         \caption{MH-SIM-FC  (Comma2k19)}
         \label{fig:mae_mh_sim_fc_comma}
     \end{subfigure}
    \begin{subfigure}[b]{0.33\textwidth}
         \centering
         \includegraphics[width=\textwidth]{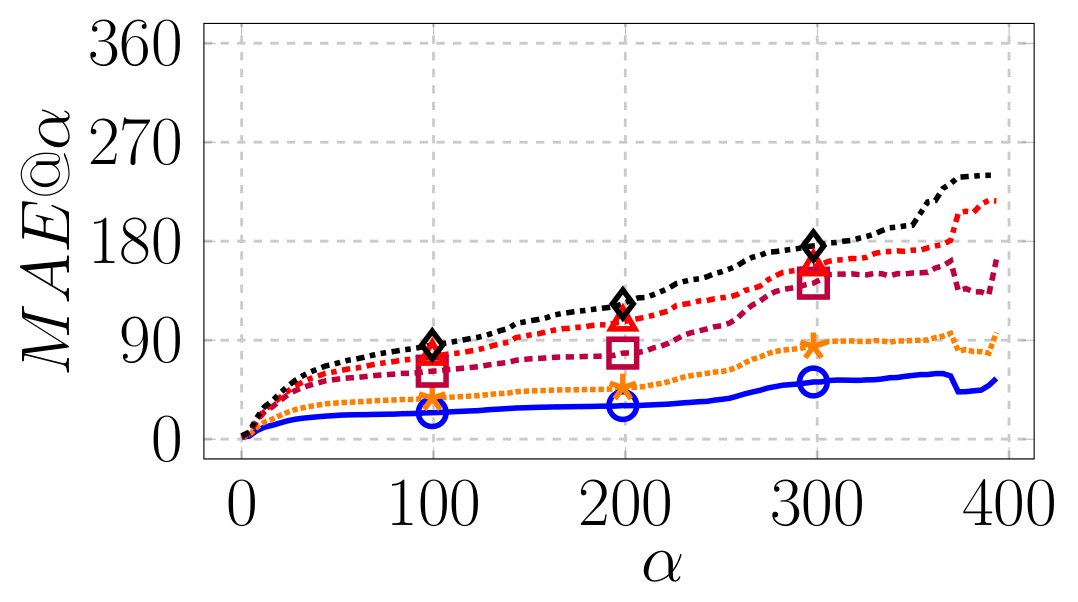}
         \caption{MH-SIM-LSTM (Comma2k19)}
         \label{fig:mae_mh_sim_lstm_comma}
     \end{subfigure}

     \begin{subfigure}[b]{0.5\textwidth}
     \centering
     \includegraphics[width=\textwidth]{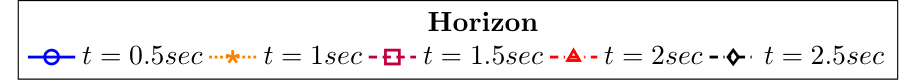}
     \end{subfigure}

    \caption{MAE@$\alpha$ for the 3 architectures for various timestamps on the two datasets. Lower values are better.}
    \label{fig:mae_graphs}
\end{figure*}

In addition, we plot the graphs of MAE@$\alpha$ as shown in Fig. \ref{fig:mae_graphs}. Notice that the error growth for MH-SIM-LSTM is slower than the error growth for both MH-IND-FC and MH-SIM-FC. Our conclusion is that MH-SIM-LSTM is more robust, and is able to deal with the imbalanced distribution of the steering angles better, which is depicted in Figure \ref{fig:steer_bias}.

\subsection{Contribution of Vision to Forecasting}
\begin{figure}[!h]
    \centering
    \begin{subfigure}[b]{0.49\linewidth}
         \centering
         \includegraphics[width=\linewidth]{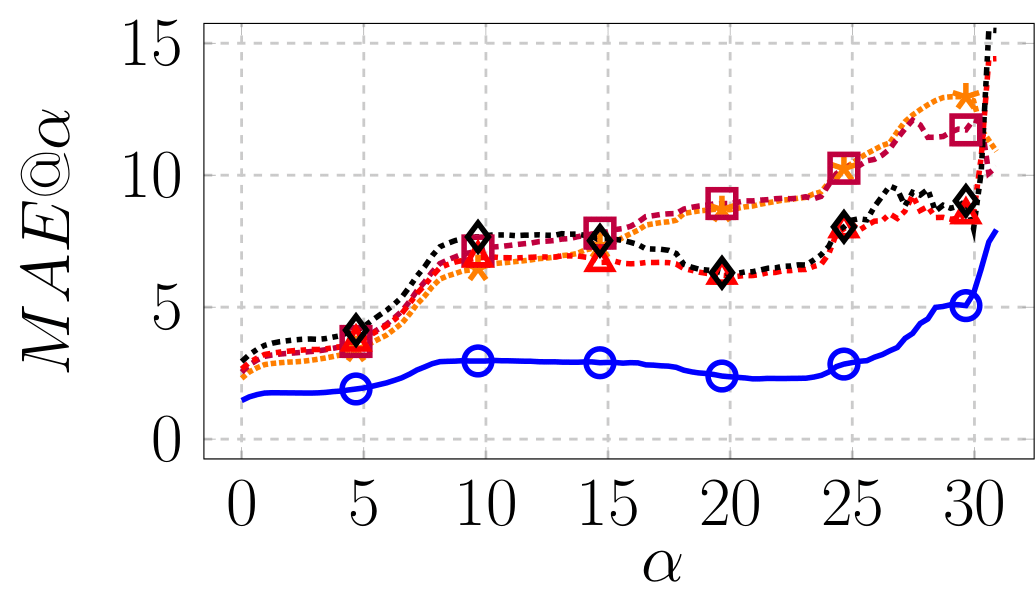}
         \caption{W/O Vision (Udacity)}
         \label{fig:mae_no_vision_udacity}
     \end{subfigure}
    \begin{subfigure}[b]{0.49\linewidth}
         \centering
         \includegraphics[width=\linewidth]{figures/mae_graphs/iccv2021_MultiHorizon_graphs/iccv2021_MultiHorizon_graphs-04.png}
         \caption{W/ Vision  (Udacity)}
         \label{fig:mae_vision_udacity}
     \end{subfigure}
     
     \begin{subfigure}[b]{0.49\linewidth}
         \centering
         \includegraphics[width=\linewidth]{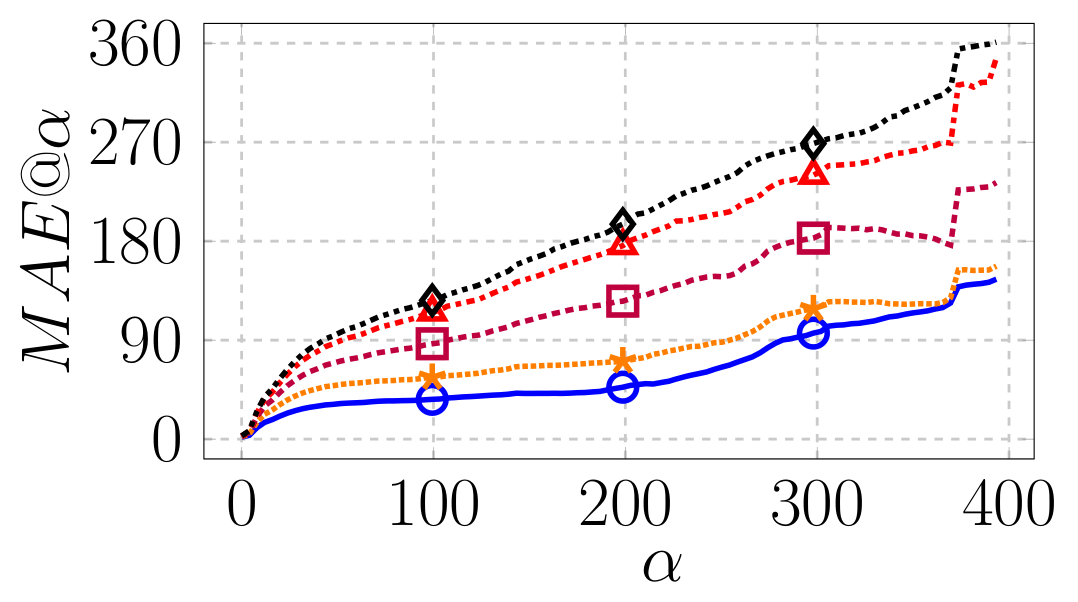}
         \caption{W/O Vision (Comma2k19)}
         \label{fig:mae_no_vision_comma}
     \end{subfigure}
    \begin{subfigure}[b]{0.49\linewidth}
         \centering
         \includegraphics[width=\linewidth]{figures/mae_graphs/iccv2021_MultiHorizon_graphs/iccv2021_MultiHorizon_graphs-11.png}
         \caption{W/ Vision  (Comma2k19)}
         \label{fig:mae_vision_comma}
     \end{subfigure}
     
     \begin{subfigure}[b]{\linewidth}
     \centering
     \includegraphics[width=\textwidth]{figures/mae_graphs/iccv2021_MultiHorizon_graphs/iccv2021_MultiHorizon_graphs-07.png}
     \end{subfigure}

    \caption{MAE@$\alpha$ for the MH-SIM-LSTM architecture, with and without using the vision stem. Lower is better. A model fed with vision achieves an error that is 56.6\% and 66.9\% of the error achieved by a model that doesn't use those features, on Udacity and Comma2k19 respectively.}
    \label{fig:mae_graphs_vision_no_vision}
\end{figure}

We experimented with multi-horizon forecasting without vision features in order to examine its effect on accuracy. Specifically, we choose the architecture that delivered the best results section \ref{section:results}, \ie, MH-SIM-LSTM and omitted the VPM stem.  The rest of the architecture remained the same. According to the results provided in Table \ref{table:vision_no_vision}, the model using only non-vision features performs significantly worse than the model using both VPM and CPM, implying that the vision features have significant positive effect on the performance. In addition, Figure \ref{fig:mae_graphs_vision_no_vision} shows the graphs for MAE@$\alpha$ for both configurations on both datasets. Notice that the error for the experiments that do not use vision input, as depicted in Figure \ref{fig:mae_no_vision_udacity} and Figure \ref{fig:mae_no_vision_comma}, the error grows dramatically faster than the experiments that use vision features, depicted in Figure \ref{fig:mae_vision_udacity} and Figure \ref{fig:mae_vision_comma}. With this, we conclude that using video clips as input for our model allows it to cope with the imbalanced distribution of the steering angles in both datasets, which is depicted in Figure \ref{fig:steer_bias}.
\begin{table}[!h]
\small
\centering
\begin{tabular}{ |c|c|c|c|c|c| }
\hline
 & & \multicolumn{2}{c|}{comma.ai} & \multicolumn{2}{c|}{Udacity} \\
 \hline
 
 Method & Horizon & MAE & RMSE & MAE & RMSE \\
 \hline
 
\multirow{5}{*}{W/ Vision} 
 & 0.5 & \textbf{1.08} & \textbf{3.444} & \textbf{0.677} & \textbf{1.394}  \\
 & 1 & \textbf{1.578} & \textbf{6.474} & \textbf{0.772} & \textbf{1.706}  \\
 & 1.5 & \textbf{1.926} & \textbf{8.378} & \textbf{0.781} & \textbf{1.733}  \\
 & 2 & \textbf{2.353} & \textbf{9.657} & \textbf{0.885} & \textbf{1.988}  \\
 & 2.5 & \textbf{2.586} & \textbf{10.442} & \textbf{1.084} & \textbf{2.235}  \\
 
 \hline
 
\multirow{5}{*}{W/O Vision} 
 & 0.5 & 1.806 & 4.689 & 1.458 & 1.458 \\
 & 1 & 2.092 & 6.718 & 2.289 & 2.289  \\
 & 1.5 & 2.506 & 8.858 & 2.528 & 2.528  \\
 & 2 & 2.805 & 10.462 & 2.64 & 2.64  \\
 & 2.5 & 3.019 & 11.562 & 2.919 & 2.919  \\
 \hline
\end{tabular}
\caption{Accuracy comparison of the architecture \emph{MH-SIM-LSTM} with the \emph{VPM} stem and without using the \emph{VPM} stem. Values are in degrees. Lower is better. In bold are the best errors for prediction for each column and horizon where is apparent the MH-SIM-LSTM with vision is the best.} 
\label{table:vision_no_vision}
\end{table}

\section{Conclusions}\label{section:discussion}
In this paper, we introduce multi-horizon forecasting of vehicle sensors, and propose a modular end-to-end deep learning based method for tackling the task. We propose and compare 3 variations of this architecture, consisting of different fusion modules and different input sources: video clips and CAN-bus data. We demonstrate through a series of extensive experiments that a methodology which exploits multi-task learning by performing multi-horizon forecasting improves the forecasting accuracy. Moreover, we show that features incorporated into video data are beneficial for forecasting of driving signals. For future work we propose two main  directions: (1) Explore more advanced architectures in order to improve both accuracy and efficiency, and (2) Experiment with more signal sources and outputs, in order to be able to forecast other useful signals.

{\small
\bibliographystyle{ieee_fullname}
\bibliography{egpaper_for_review}
}

\end{document}